\documentclass[runningheads]{llncs}
\usepackage{graphicx}
\usepackage{comment}
\usepackage{amsmath,amssymb} % define this before the line numbering.
\usepackage{color}
\usepackage{times}
\usepackage{epsfig}
\usepackage{multirow}
\usepackage{multicol}
\usepackage{booktabs}
\usepackage{array}
\usepackage{threeparttable}
\usepackage{physics}
\usepackage{bm}
\usepackage{fixltx2e}
\usepackage{dblfloatfix}
\usepackage[colorlinks,linkcolor=red]{hyperref}

\usepackage{floatrow}
\usepackage[width=122mm,left=12mm,paperwidth=146mm,height=193mm,top=12mm,paperheight=217mm]{geometry}

\usepackage{caption}
\usepackage{subcaption}
\begin{document}
% \linenumbers
\pagestyle{headings}
\mainmatter
\def\ECCVSubNumber{5996}  % Insert your submission number here

\title{Rethinking of Pedestrian Attribute Recognition: Realistic Datasets and A Strong Baseline} % Replace with your title

\begin{comment}
%\titlerunning{ECCV-20 submission ID \ECCVSubNumber} 
%\authorrunning{ECCV-20 submission ID \ECCVSubNumber} 
%\author{Anonymous ECCV submission}
%\institute{Paper ID \ECCVSubNumber}
\end{comment}
%******************

%\begin{comment}
\titlerunning{Rethinking of Pedestrian Attribute Recognition}

\author{Jian Jia\inst{1,2} \and Houjing Huang\inst{1,2} \and Wenjie Yang\inst{1,2} \and Xiaotang Chen\inst{1,2} \and Kaiqi Huang\inst{1,2,3}}

%\author{First Author\inst{1} \and Second Author\inst{2,3} \and
%Third Author\inst{3}}

\authorrunning{Jian Jia et al.}
\institute{School of Artificial Intelligence, University of Chinese Academy of Sciences, Beijing, China \and
CRISE, Institute of Automation, Chinese Academy of Sciences, Beijing, China \and 
CAS Center for Excellence in Brain Science and Intelligence Technology, Beijing, China
{jiajian2018@ia.ac.cn, \{houjing.huang,wenjie.yang,xtchen,kaiqi.huang\} @nlpr.ia.ac.cn}
}
%\end{comment}
%******************
\maketitle

\begin{abstract}
Despite various methods are proposed to make progress in pedestrian attribute recognition, a crucial problem on existing datasets is often neglected, namely, a large number of identical pedestrian identities in train and test set, which is not consistent with practical application. Thus, images of the same pedestrian identity in train set and test set are extremely similar, leading to overestimated performance of state-of-the-art methods on existing datasets. To address this problem, we propose two realistic datasets PETA\textsubscript{$zs$} and RAPv2\textsubscript{$zs$} following zero-shot setting of pedestrian identities based on PETA and RAPv2 datasets. Furthermore, compared to our strong baseline method, we have observed that recent state-of-the-art methods can not make performance improvement on PETA, RAPv2, PETA\textsubscript{$zs$} and RAPv2\textsubscript{$zs$}. Experiments on existing and proposed datasets verify the superiority of our method by achieving state-of-the-art performance. Code is available at \url{https://github.com/valencebond/Strong_Baseline_of_Pedestrian_Attribute_Recognition}

%\dots
\end{abstract}

\section{Introduction}
Pedestrian attribute recognition \cite{zhu2013pedestrian} is to predict multiple attributes of pedestrian images as semantic descriptions in video surveillance, such as age, gender and clothing. 

Recently, pedestrian attribute recognition has drawn increasing attention due to its great potential in real world application such as person retrieval \cite{li2018richly}, person search \cite{feris2014attribute} and person re-identification \cite{yang2019towards,lin2019improving}. Similar as many vision tasks, progress on pedestrian attribute recognition is significantly advanced by deep learning. From the pioneer work DeepMAR \cite{li2015deepmar} based on CaffeNet \cite{donahue2014decaf} to most recent work VAC \cite{guo2019visual} based on ResNet50 \cite{he2016deep}, mA performance is increased from 73.79 to 78.47 in RAPv1 \cite{li2016richly} dataset \footnote{We use RAPv1, RAPv2 to represent dataset published by Li et al. \cite{li2016richly} and Li et al. \cite{li2018richly} respectively.}. 
 \begin{figure}
	\centering
	\begin{subfigure}[b]{0.26\linewidth}
         \centering
         \includegraphics[width=\linewidth]{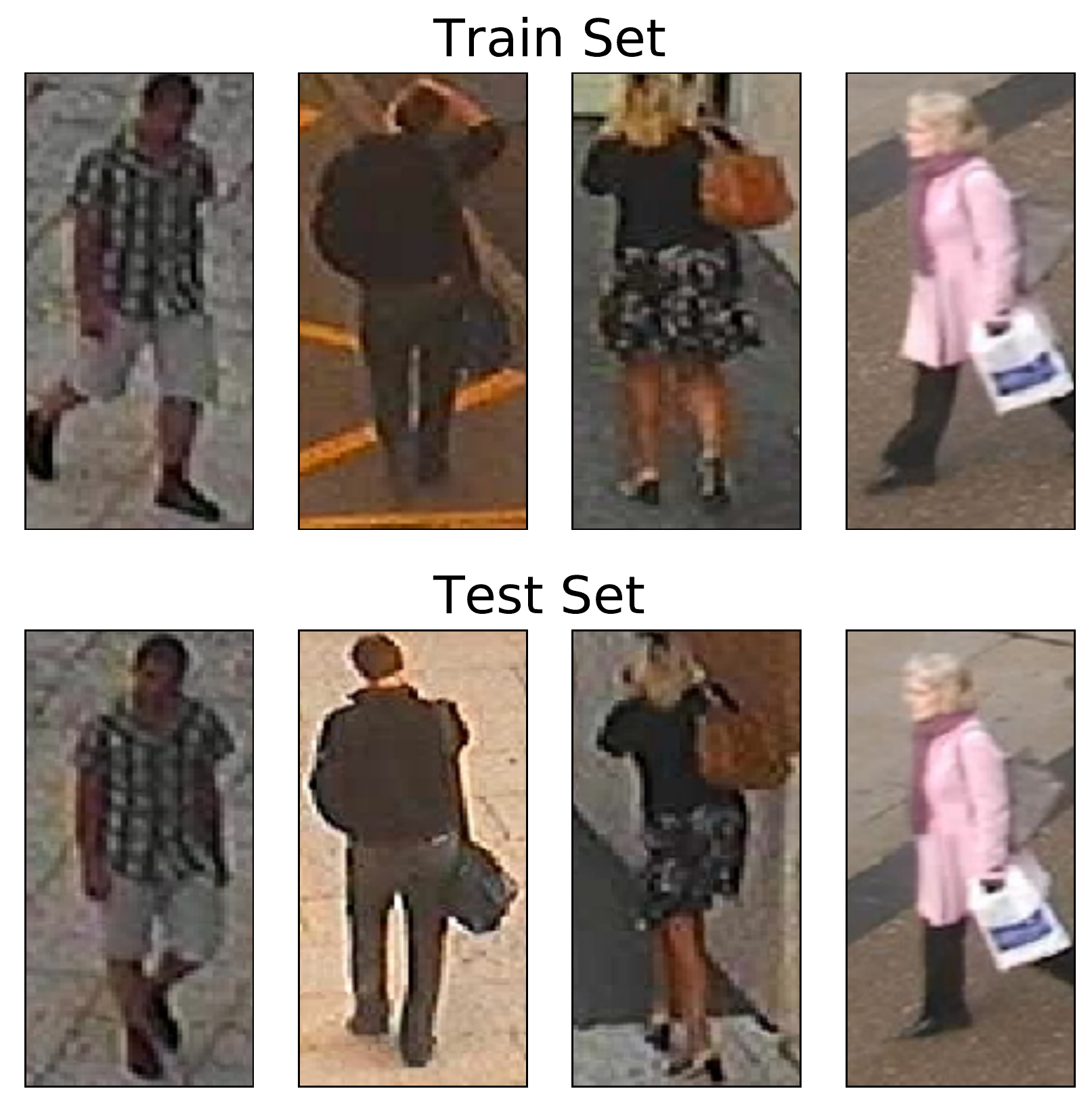}
         \caption{PETA dataset}
         \label{fig:overlap_id1}
         
     \end{subfigure}
     \hfill
     \begin{subfigure}[b]{0.26\linewidth}
         \centering
         \includegraphics[width=\linewidth]{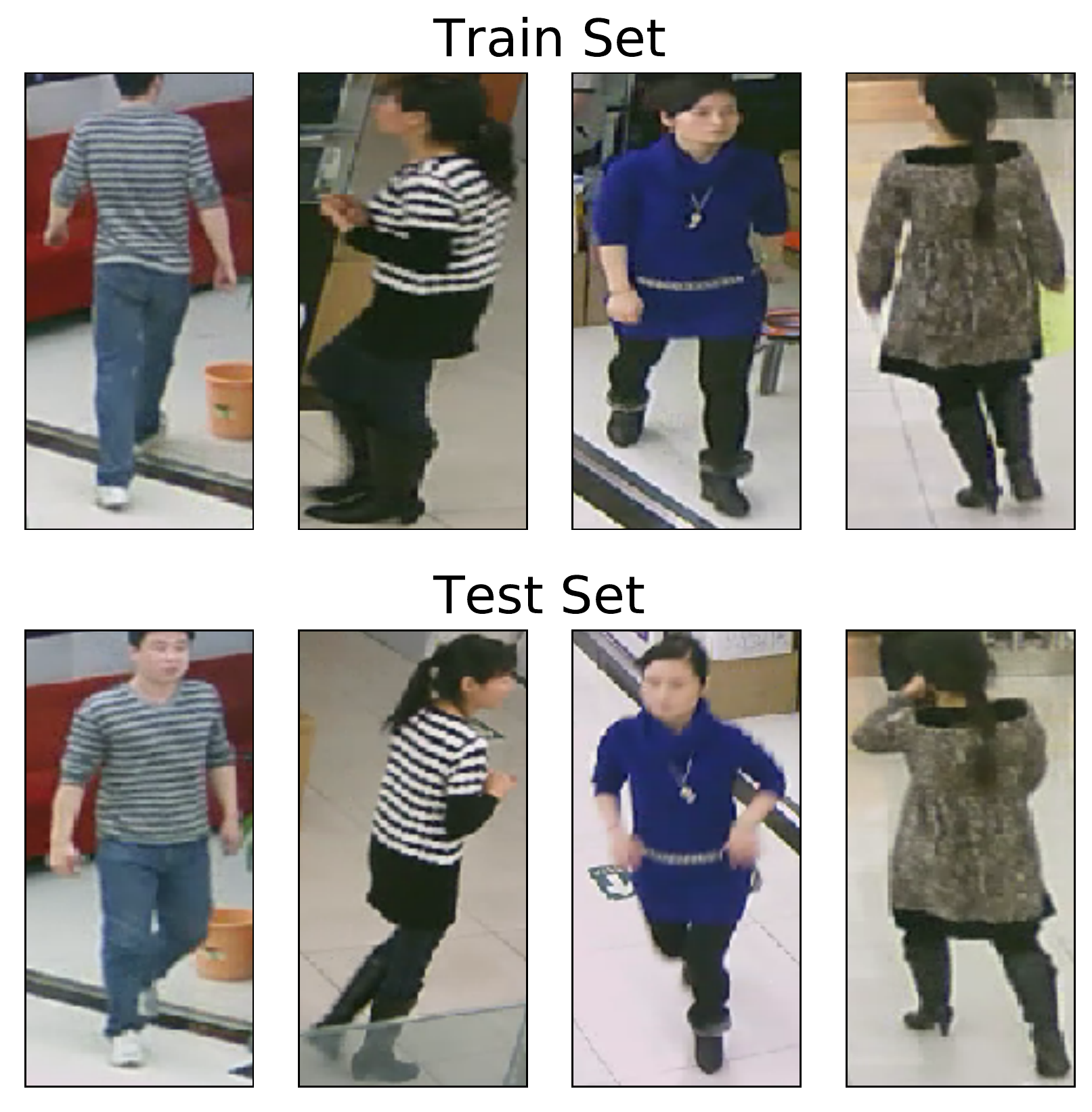}
         \caption{RAPv2 dataset}
         \label{fig:overlap_id2}
     \end{subfigure}
     \hfil
     \begin{subfigure}[b]{0.46\linewidth}
         \centering
         \includegraphics[width=\linewidth]{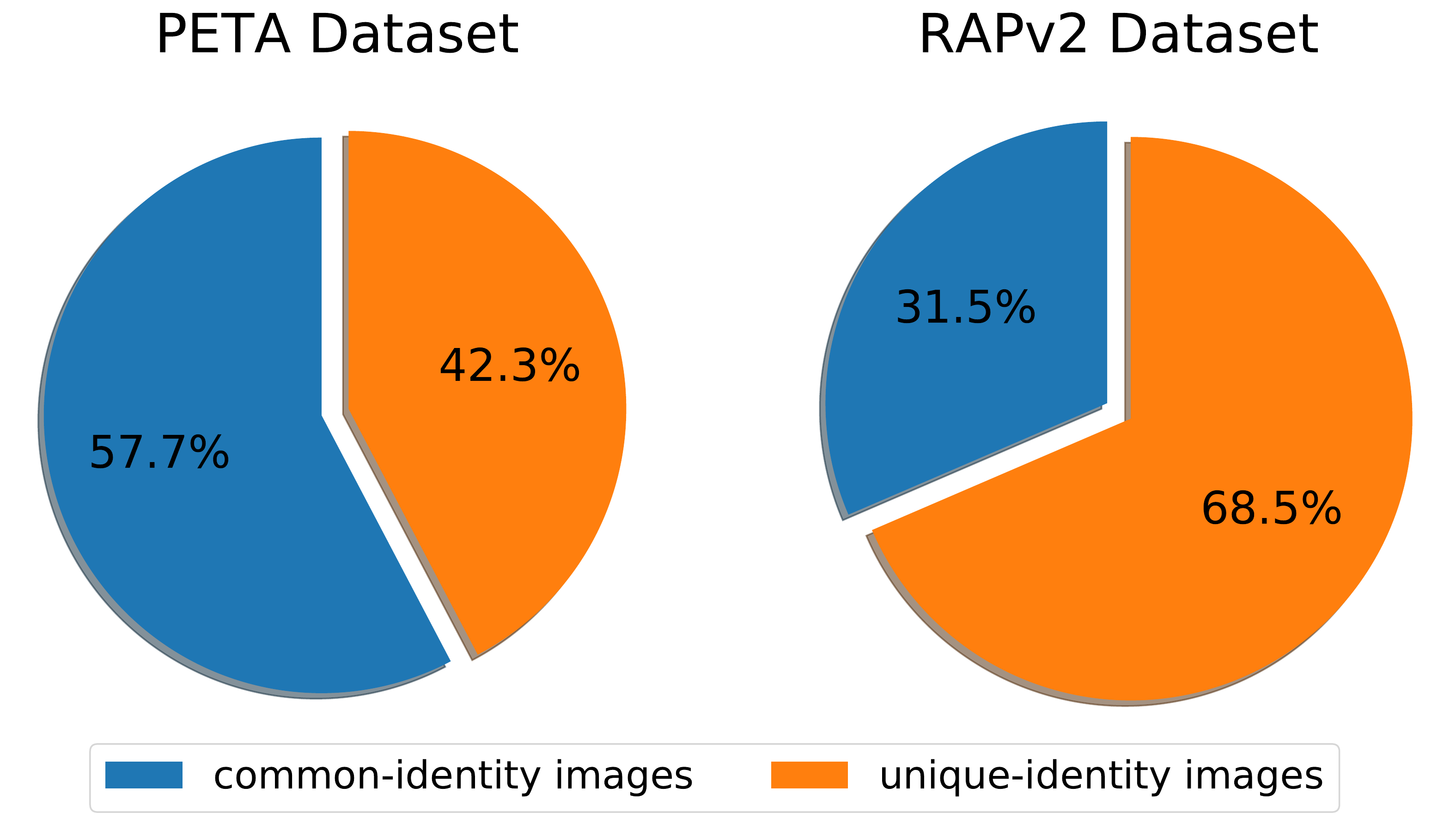}
         \caption{Overlap Proportion}
         \label{fig:overlap_id3}
     \end{subfigure}
     
     \caption{Extremely similar images of the same pedestrian identity in train set and test set. (a) Images in PETA dataset. (b) Images in RAPv2 dataset. (c) The proportions of common-identities images in the test sets of PETA and RAPv2. The proportion of common-identity images in RAPv2 test set is at least 31.5\%, due to some images are not labeled with pedestrian identity.}
     \label{fig:overlap_id}
\end{figure}

However, when various networks are proposed to improve the performance by extracting more discriminative features, a crucial problem in existing popular datasets is often neglected, \textit{i.e.} there are a large number of identical pedestrian identities in train and test set. This results in plenty of extremely similar images of the same pedestrian identity in the train and test set. For example, in PETA \cite{deng2014pedestrian} and RAPv2 \cite{li2018richly}, the first large-scale pedestrian attribute dataset and the updated version of the most popular dataset RAPv1 \cite{li2016richly}, the images of the same pedestrian identity in train set and test set are almost the same except for negligible background and pose variation, as shown in Fig.~\ref{fig:overlap_id}\subref{fig:overlap_id1} and Fig.~\ref{fig:overlap_id}\subref{fig:overlap_id2}.

The proportion of common-identity \footnote{\textit{Common-identity} indicates pedestrian identity exists both in train set and test set. \textit{Unique-identity} indicates the identity only exists in train set or test set.} images in test set are calculated as shown in Fig.~\ref{fig:overlap_id}\subref{fig:overlap_id3}. It is worth noting that 57.5\% and 31.5\% of test set images have similar counterparts of the same pedestrian in the train set of PETA and RAPv2 respectively. Although there are also plenty of similar images in train set and test set of RAPv1, we can not get the accurate proportion of common-identity images in test set, due to pedestrian identity label is not provided. Thus, existing datasets are unreasonable in practical application in which test set identities barely have overlap with identities of train set. 

\begin{figure}[h]
	\centering
	\begin{subfigure}[b]{0.49\linewidth}
         \centering
         \includegraphics[width=\linewidth]{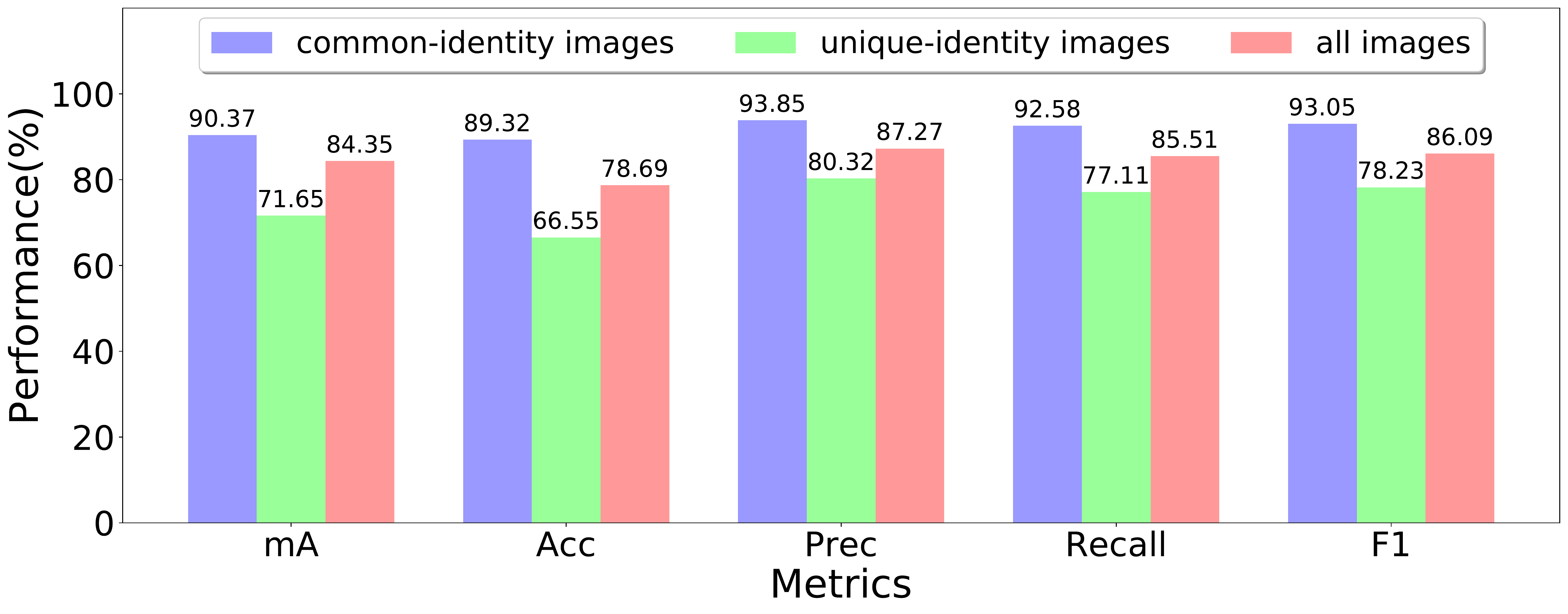}
         \caption{MsVAA}
         \label{fig:overlap_id1}
     \end{subfigure}
     \hfill
     \begin{subfigure}[b]{0.49\linewidth}
         \centering
         \includegraphics[width=\linewidth]{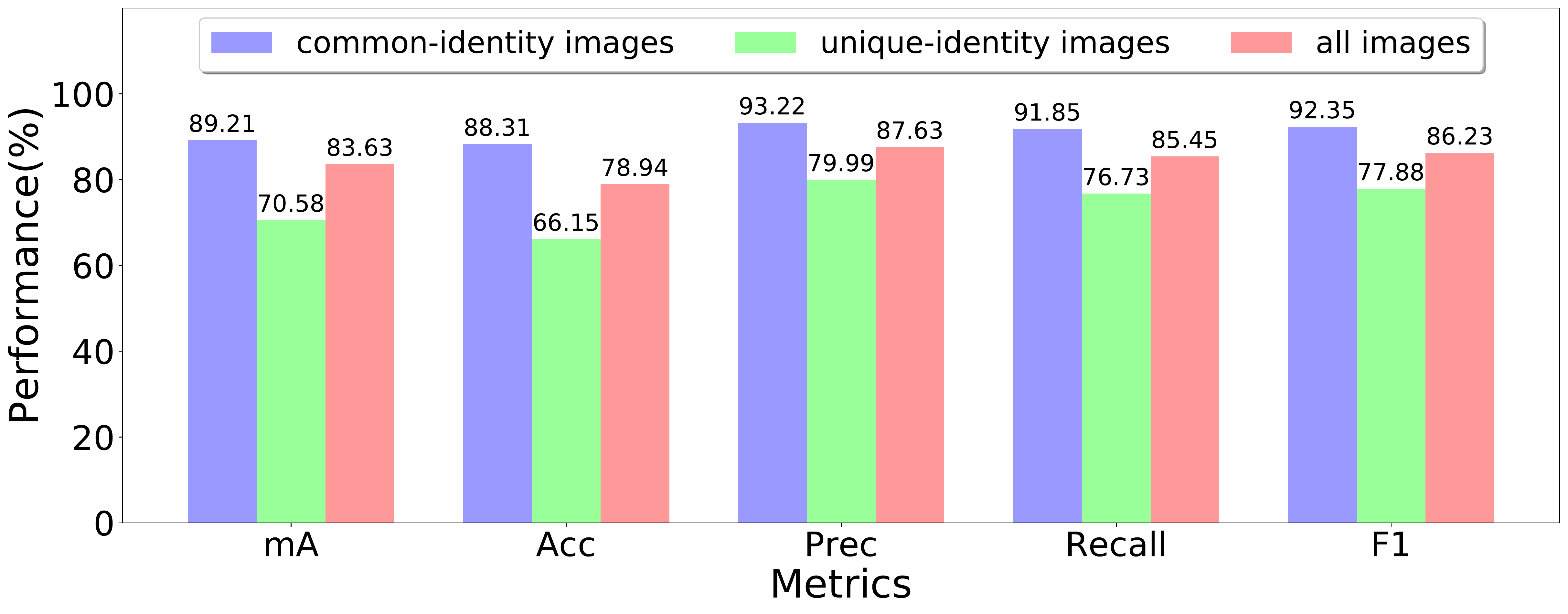}
         \caption{VAC}
         \label{fig:overlap_id2}
     \end{subfigure}
     \caption{Performance of MsVAA \cite{sarafianos2018deep} and VAC \cite{guo2019visual} methods on common-identity images, unique-identity images and all images of PETA test set. There is a significant performance gap between common-identity images and unique-identity images as well as unique-identity images and all images of test set. The remarkable performance gap shows the irrationality of existing datasets. Similar phenomenon can be observed on ALM \cite{tang2019Improving} method and RAPv2 dataset as well.}
     \label{fig:overlapped_img_gap_msvaa}
     \vspace{-.5em}
\end{figure}

More importantly, the performance of state-of-the-art(SOTA) methods is overestimated on existing datasets, which misleads the progress of pedestrian attribute recognition. To verify our hypothesis, we reimplement the recent SOTA methods MsVAA \cite{sarafianos2018deep}, VAC \cite{guo2019visual}, ALM \cite{tang2019Improving} and experiments are conducted to evaluate the performance of common-identity images and unique-identity images separately on PETA. As illustrated in Fig.~\ref{fig:overlapped_img_gap_msvaa}(a), the performance gaps of MsVAA between common-identity images and unique-identity images are 20.51\%, 24.70\%, 15.12\%, 16.87\%, 16.32\% in mA, Acc, Precision, Recall and F1 respectively. Significant performance gaps validates the existing datasets are impractical under real scenario. And compared to the performance of all the images in the test set, the performance of unique-identity images is reduced by 12.7\%, 12.14\%, 6.95\%, 8.4\%, 7.86\% respectively in mA, Acc, Precision, Recall and F1, which proves that the performance of existing methods is overestimated. Similar performance degradation is also observed for VAC and ALM methods on RAPv2 as well.

To address the problem in existing datasets, we propose a realistic  setting for pedestrian attribute datasets: pedestrian identities of test set have no overlap with identities of train set, i.e. pedestrian identities follow the zero-shot setting. Based on PETA \cite{deng2014pedestrian} and RAPv2 \cite{li2018richly} datasets provided with pedestrian identity label, we construct two realistic datasets PETA\textsubscript{$zs$} and RAPv2\textsubscript{$zs$} to make them in line with zero-shot setting of pedestrian identities illustrated in Fig.~\ref{fig:overlap_dataset}. Consistent performance drop of different SOTA methods in proposed datasets highlights the rationality of our proposed datasets as detailed in Table~\ref{tab:proposed_dataset_perf} .

Furthermore, we find experimentally that SOTA methods can not make performance improvement based on our strong baseline method. There are two reasons. First, performance improvement gained by recent SOTA methods is based on underutilized baseline. Second, SOTA methods resort to localize the attribute-specific region to achieve better performance by attention module \cite{sarafianos2018deep,guo2019visual} or Spatial Transformer Network(STN) \cite{tang2019Improving}. However, the strong baseline itself can achieve a good localization of the attribute area, so further strengthening the localization capability cannot bring more performance improvements.

The contributions of this paper are as follows:
 
\begin{itemize}
	\item We observe the crucial problem in existing pedestrian attribute datasets, i.e. a large number of identical pedestrian identities in train and test set, which is impractical and misleads the model evaluation.
	\item To solve the dataset problem, we propose two datasets PETA\textsubscript{$zs$} and RAPv2\textsubscript{$zs$} following zero-shot setting of pedestrian identities.
	\item Based on our strong baseline, we experimentally find that  enhancing localization of attribute-specific area adopted by SOTA methods is not beneficial for performance improvement. 
\end{itemize}

The rest of this paper is organized as follows. Section \ref{related_work} reviews the related work of pedestrian attribute recognition. Section \ref{dataset} rethinks existing pedestrian attribute setting and proposes two realistic datasets. Section \ref{method} proposes our method  and experiments are conducts on Section \ref{exp}. Finally, section \ref{conclusion} concludes this work and discusses future directions. 

\section{Related Work} \label{related_work}

\subsection{Pedestrian Attribute Recognition}
Most recent efforts resort to learning discriminate attribute-specific feature representation by enhancing attribute localization.
Li \textit{et al.} \cite{li2015deepmar} firstly considered pedestrian attribute recognition as a multi-label classification task and proposed the weighted sigmoid cross entropy loss. 
Considering better attribute localization can reduce the impact of irrelevant areas, Liu \textit{et al.} \cite{liu2017hydraplus} proposed HydraPlus-Net with multi-directional attention modules to locate fine-grained attributes. 
Liu \textit{et al.} \cite{liu2018localization} proposed a Localization Guided Network based on Class Activation Map(CAM) \cite{zhou2016learning} and EdgeBox \cite{zitnick2014edge} to extract attribute-specific local features. 
Considering corresponding area scales of attributes are various, multi-scale feature maps were reused from different backbone layers \cite{yu2016weakly,sarafianos2018deep}, \textit{i.e.} pyramidal feature hierarchy, instead of single feature map of last layer. 
Guo \textit{et al.} \cite{guo2019visual} utilized the assumption that visual attention regions are consistent between different spatial transforms of same image and proposed an attention consistency loss to get a robust attribute localization. 
Inspired by Feature Pyramid Network(FPN) \cite{lin2017feature}, Tang \textit{et al.} \cite{tang2019Improving} constructed Attribute Localization Module with Squeeze-and-Excitation(SE) block \cite{hu2018squeeze} and Spatial Transformer Network (STN) \cite{jaderberg2015spatial} to enhance attribute localization. 
From the viewpoint of capturing attribute semantic dependencies, some methods focused on modeling attribute relationship. Wang \textit{et al.} \cite{wang2017attribute} transformed pedestrian attribute recognition to sequence prediction by Long-Shot-Term-Memory (LSTM) \cite{hochreiter1997long} to explore attribute context and correlation.

Compared to previous work, our work rethinks recent progress made on pedestrian attribute recognition from the perspective of datasets and methods. First, the dataset problem we first noticed leads to overestimated performance and misleads the evaluation of recent methods. Thus, we propose two reasonable and realistic datasets. Second, based on a fully utilized baseline network, we find localizing specific-attribute area adopted by recent SOTA methods \cite{sarafianos2018deep,guo2019visual,tang2019Improving} is not beneficial for performance improvement.

\section{Proposed Realistic Datasets} \label{dataset}

In this section, we first answer two questions: what's wrong with the existing pedestrian attribute datasets and Why is it important for academic research and industry applications. Then we introduce a new realistic setting and propose two reasonable and practical datasets PETA\textsubscript{$zs$}, RAPv2\textsubscript{$zs$}.

\subsection{Problems of existing Datasets}

As far as we know, APiS \cite{zhu2013pedestrian} is the first pedestrian attribute recognition dataset followed by PETA \cite{deng2014pedestrian}, RAPv1 \cite{li2016richly}, PA100k \cite{liu2017hydraplus}, RAPv2 \cite{li2018richly} which promote the development of pedestrian attribute recognition. However, no clear, concrete and unified setting is proposed for pedestrian attribute dataset construction. 

For PETA, RAPv1, RAPv2 datasets, randomly splitting is adopted as default setting to construct train and test set and the protocol are used by almost all methods \cite{li2016richly,liu2017hydraplus,wang2017attribute,li2018richly,sarafianos2018deep,guo2019visual,tang2019Improving}. This results in a large number of identical pedestrian identities in train and test set.
Considering images of identical pedestrian identities are often collected by the same surveillance camera in very short frame intervals, the appearance of these images is extremely similar with negligible background and pose variation as detailed in Fig.~\ref{fig:overlap_id}(a) and Fig.~\ref{fig:overlap_id}(b).
As a result, there are plenty of extremely similar images in train set and test set as shown in Fig.~\ref{fig:overlap_id}(c).

\begin{figure}[h]
\centering
\includegraphics[width=0.9\linewidth]{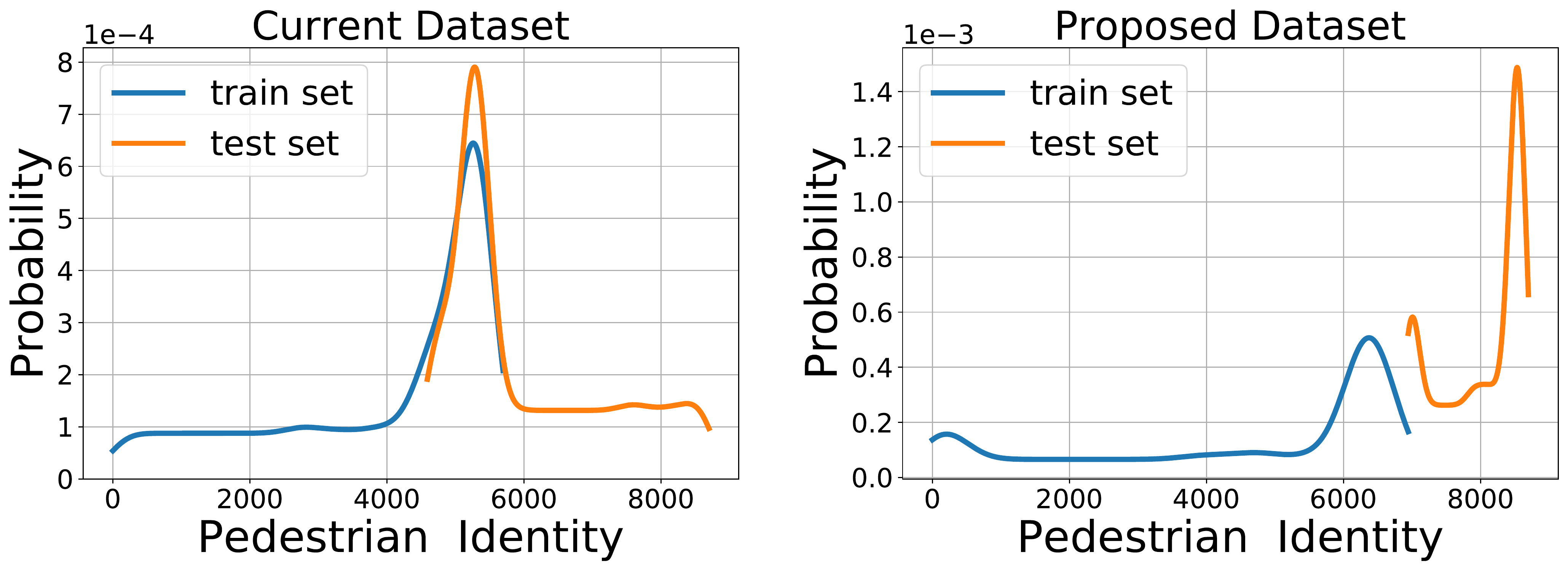}
%\vspace{-0.5em}
\caption{Pedestrian identity distribution on PETA and PETA\textsubscript{$zs$}. X-axis indicates the pedestrian identity number and Y-axis indicates the proportion of corresponding images of the pedestrian identity. There are 1106 common identities in train and test set, accounting for 19.42\% identities (55.27\% images) of train set and 26.91\% identities (57.70\% images) of test set. Our proposed dataset PETA\textsubscript{$zs$} solves the problem by completely separating the identities of test set from the identities of train set.}
\label{fig:overlap_dataset}
\end{figure}

To further illuminate the difference between existing datasets and our proposed datasets, the pedestrian identity distribution of PETA is given in Fig.~\ref{fig:overlap_dataset}. In existing dataset PETA, there are 1106 identical identities in train and test set, accounting for 26.91\% identities and 57.70\% images of the test set. It means that more than half of test set images have their similar counterparts in train set, which is similar to 'data leakage'. In our proposed PETA\textsubscript{$zs$}, there is no overlap between pedestrian identities of train and test set. As for RAPv1 (or RAPv2), all images (or part of images) have no pedestrian identity annotation, so we can not get the accurate identity distribution.

Given the problem of existing datasets, the reasons why it is crucial for industry application and academic research are given as follows. Whether used as primary task in video surveillance or auxiliary task in person retrieval, pedestrian identities of test set barely overlap with identities of train set. So, existing datasets setting is inconsistent with real world application. 
More importantly, the performance of SOTA methods is overestimated on existing datasets and the model evaluation is misled. 
We reimplement recent methods MsVAA \cite{sarafianos2018deep}, VAC \cite{guo2019visual}, ALM \cite{tang2019Improving} and report their performance in Fig.~\ref{fig:overlapped_img_gap_msvaa}. Compared to performance on the whole test set, consistent performance degradation of three methods on unique-identity images of test set confirms the overestimated model performance.

\subsection{Construction of Proposed Datasets}

To solve the problem of existing pedestrian attribute datasets, we propose a realistic setting: there is no identical pedestrian identities in train and test set, \textit{i.e.} pedestrian identities follow the zero-shot setting. Concretely, we propose the following criteria for dataset construction, to serve as reference for future dataset.
\begin{description}
    \item[Criteria of proposed datasets:]
\end{description}
\begin{center}
  \begin{minipage}{0.9\linewidth}
    \begin{enumerate}
    \item $\mathbb{I}_{all} =\mathbb{I}_{train} \cup \mathbb{I}_{valid} \cup \mathbb{I}_{test}$, \\ $\abs{\mathbb{I}_{train}} :\abs{\mathbb{I}_{valid}} : \abs{\mathbb{I}_{test}}$ $\approx$  $ 3 : 1 : 1 $ .
    \item $\mathbb{I}_{train} \cap \mathbb{I}_{valid} = \varnothing$, \\$ \mathbb{I}_{train} \cap \mathbb{I}_{test} = \varnothing $, $\mathbb{I}_{valid} \cap \mathbb{I}_{test} = \varnothing $ .
    \item $\abs{ \abs{\mathbb{I}_{valid}} - \abs{\mathbb{I}_{test}} } < \abs{\mathbb{I}_{all}} \times  T_{id}$ .
%    \item $\abs{\abs{I_{valid}} - \abs{I_{test}}} < T_{id} \times (\abs{I_{train}} + \abs{I_{valid}} + \abs{I_{test}})$ .
    \item $\abs{N_{valid} - N_{test}} < T_{img}$ .
    \item $\abs{R_{valid} - R_{test}} < T_{attr}$ .
    \end{enumerate}
  \end{minipage}
\end{center}
where $\mathbb{I}$ denote the pedestrian identity set, $N$, $R$, $T$ denote the number of images, the ratio of positive samples of attributes and pre-defined threshold separately. $T_{id} = 0.01$, $T_{img} = 300$, $T_{attr} = 0.03$ are used in our experiments by default. Subscript ${train}$, ${valid}$, ${test}$ denotes train set, validation set and test set separately. $\abs{\ \cdot \ }$ is the set cardinality.

To solve the problem of identical identities between train set and test set, \textbf{Criterion 1,2,3} are proposed to repartition train set and test set of PETA and RAPv2 datasets based on pedestrian identities to make sure that pedestrian identities follow zero-shot setting. To better evaluate the model performance and control the attribute distribution difference between train set and test set, \textbf{Criterion 4, 5} are proposed.

Based on criteria mentioned above, considering pedestrian identity label is only provided in PETA and RAPv2 datasets, two realistic datasets PETA\textsubscript{$zs$} and RAPv2\textsubscript{$zs$} are proposed, where subscript $zs$ denotes pedestrian identities of dataset following zero-shot setting. 
%For PETA dataset, we divide all 19,000 images into train set, validation set and test set following the criteria. And identity distribution of  proposed PETA\textsubscript{$zs$} dataset are illustrated in the right of Fig.~\ref{fig:overlap_dataset}, where identities of train set and test set are completely separated and have no overlap.
%
%For RAPv2 dataset, 84,928 pedestrian images consist of 26,638 images labeled with pedestrian identity, 43,343 images without pedestrian identity and 14,947 distractor images. Thus, we abandon the images without pedestrian identity label and distractor images, and only divide identity-labeled 26,638 images into train set, valid set and test set according to the criteria. 
%For PA100k dataset, pedestrian identities of test set have no overlap with identities of train set. So we keep the dataset setting unchanged. For RAPv1 dataset, pedestrian identity labels are not provided. So we leave it alone.
Details of the two proposed dataset are given in Table~\ref{tab:dataset_partition} and supplementary material.

\section{Methods} \label{method}

\subsection{The Strong Baseline}
Given a dataset $\mathbb{D}=\{(\bm{X}_i, \bm{y}_i)\}$, $\bm{y}_i \in \{0, 1\}^{M}$, $i=1, 2, ..., N$, where $\bm{y}_i$ indicates ground truth vector and $N$, $M$ denotes the number of train images and attributes respectively. $\bm{X}_i$ denotes $i$-$th$ pedestrian image. Pedestrian attribute recognition is a multi-label task to learn to predict attributes $\bm{\hat{y}}_i \in \{0, 1\}^{M}$, given the pedestrian image $X_i$. The element values of zeros and ones in the label vector $\bm{y}$ and $\bm{\hat{y}}$ denote the absence and presence of the corresponding attributes in the pedestrian image.

Pedestrian attribute model generally adopts multiple binary classifiers with sigmoid function \cite{guo2019visual,tang2019Improving} instead of a multi-class classifier with softmax function \cite{wang2017normface,deng2019arcface}. So the loss can be computed as 
\begin{align}
	Loss & = \frac{1}{N} \sum_{i=1}^{N} \sum_{j=1}^{M} \omega_{j}(y_{ij} \log(\sigma(logits_{ij})) + (1 - y_{ij})\log(1 - \sigma(logits_{ij}))) 
	\label{eq:celoss}
\end{align}
where $\sigma(z) = 1/(1+ e^{-z})$ and $logits_{ij}$ is the output of classifier layer and $\omega_{j}$ adopted here is proposed in \cite{li2018pose} to alleviate the distribution imbalance between attributes.
\begin{align}
	\omega_{j} = 
    \begin{cases}
      e^{1 - r_{j}}, &  y_{ij}=1 \\
      e^{r_{j}}, & y_{ij}=0
    \end{cases}
\end{align}
where $r_{j} $ is the positive sample ratio of $j$-$th$ attribute in train set. 

%and the corresponding label vector as $\bm{y}_i \in \{0, 1\}^{M}$ 

Denote the feature representation of the $i$-$th$ example as $\mathbf{x}_i \in R^{d}$ , then the conditional probability output by a deep neural network can be predicted by a linear classifier followed by a sigmoid function in Eq.~(\ref{eq:linear_classifier})
\begin{align}
	p_{ij} = Pr(Y=y_{ij}|\mathbf{x}_i) = \frac{1}{1 + e^{{-\mathbf{w}^{T}_{j} \mathbf{x_i}}}}
	\label{eq:linear_classifier}
\end{align}
where $[\mathbf{w}_1, \mathbf{w}_2, ... , \mathbf{w}_M] \in R^{d \times M }$ is weight of linear classifier and $p_{ij}$ is the $j$-$th$ attribute probability of the $i$-$th$ image. We denote this method as baseline in the following sections.

\section{Experiments} \label{exp}

\subsection{Datasets and Evaluation Metrics}

We conduct experiments in four existing pedestrian attribute datasets, PETA \cite{deng2014pedestrian}, RAPv1 \cite{li2016richly}, RAPv2 \cite{li2018richly}, PA100k \cite{liu2017hydraplus} and two proposed realistic datasets PETA\textsubscript{$zs$}, RAPv2\textsubscript{$zs$}. Details of each dataset are given in Table~\ref{tab:dataset_partition}.

\begin{table}[ht]
    \centering
        \caption{Details of existing datasets and proposed datasets. Zero-shot setting of pedestrian identities is considered in proposed PETA\textsubscript{$zs$} and RAPv2\textsubscript{$zs$} datasets. $I_{train}$, $I_{valid}$, $I_{test}$ indicates the number of identities in train set, validation set and test set respectively. Pedestrian identities are not provided in RAPv1, PA100k and are partly provided in RAPv2, so the exact quantity cannot be counted, which is denoted by -- . Due to the overlapped identities between $I_{train}$, $I_{valid}$ and $I_{test}$ of PETA, the sum of $I_{train}$, $I_{valid}$ and $I_{test}$ is not equal to that in PETA\textsubscript{$zs$}. \textit{Attribute} here denotes the number of attributes used for evaluation.}
    \label{tab:dataset_partition}

    \begin{tabular}{c|c|c|c|c|c|c|c|c|c}
    \toprule
        Dataset & Setting & $I_{train}$ & $I_{valid}$ & $I_{test}$  & Attribute & Images & $N_{train}$ & $N_{val}$ & $N_{test}$ \\ \hline \hline
        PETA & existing & 4,886 & 1,264  & 4,110  & 35 & 19,000 & 9,500 & 1,900 & 7,600 \\
        PETA\textsubscript{$zs$} & zero-shot & 5,211 & 1,703 & 1,785 & 35 & 19,000 & 11,051 & 3,980 & 3,969 \\
        RAPv2 & existing & -- & -- & --  & 54 & 84,928 & 50,957 & 16,986 & 16,985 \\
        RAPv2\textsubscript{$zs$} & zero-shot & 1,508 & 546 & 535  & 54 & 26,632 & 14,729 & 5,961 & 5,948 \\
        RAPv1 & existing & -- & -- & -- & 51 & 41,585 & 33,268 & -- & 8,317 \\
        PA100k & existing & -- & -- & -- & 26 & 100,000 & 80,000 & 10,000 & 10,000  \\
    \bottomrule
    \end{tabular}
    \vspace{-2em}
\end{table}

Two types of metrics, \textit{i.e.} a label based metric and four instance based metrics, are adopted to evaluate attribute recognition performance \cite{li2018richly}. For label based metric, we compute the mean value of classification accuracy of positive samples and negative samples as the metric of specific attribute, then we take an average over all attributes as mean Accuracy ({\bf mA}). For instance based metrics, {\bf Accuracy}, {\bf Precision}, {\bf Recall} and {\bf F1-score} are used.

\subsection{Implementation Details}
The proposed method is implemented with PyTorch and trained end-to-end. ResNet50 \cite{he2016deep} is adopted as backbone to extract pedestrian image feature. Pedestrian images are resized to $256 \times 192$ with random horizontal mirroring as inputs. SGD is employed for training, with momentum of 0.9, and weight decay of 0.0005. The initial learning rate equals to 0.01 and batch size is set to 64. Plateau learning rate scheduler is used with reduction factor 0.1. Total epoch number of training is 30

\subsection{Comparison with State-of-the-art Methods}

\noindent {\bf Results on proposed datasets} To fully validate the rationality of proposed datasets, we reimplement the recent methods MsVAA \cite{sarafianos2018deep}, VAC \cite{guo2019visual} and ALM \cite{tang2019Improving} as MsVAA\footnotemark[1], VAC\footnotemark[1] and ALM\footnotemark[1] respectively based on backbone ResNet50 \cite{he2016deep} . Quantitative experiments are conducted in PETA\textsubscript{$zs$}, RAPv2\textsubscript{$zs$} datasets and results are reported in Table~\ref{tab:proposed_dataset_perf}. 
It is worth noting that, compared to existing datasets, remarkable performance drop of all methods exists on proposed datasets even if PETA\textsubscript{$zs$} have more train images compared to PETA as shown in Table~\ref{tab:dataset_partition}. The result further validates our insight that performance on existing datasets is overestimated and existing datasets mislead the model evaluation. We find experimentally that there is a trade-off between Precision and Recall, so mA and F1 score are more reliable and convinced.The proposed method achieves state-of-the-art performance, with significantly less parameters and computation.
%\vspace{-0.5em}
\begin{table}[ht]
\vspace{-0.5em}
\centering
\caption{Performance comparison of four methods on the PETA, RAPv2 datasets. We use \textit{zero-shot} setting to denote our proposed PETA\textsubscript{$zs$} or RAPv2\textsubscript{$zs$} dataset. Five metrics, mA, Accuracy, Precision, Recall, F1 are evaluated. Parameters(Params) and multiply-accumulate operations(MACs) of various methods are also reported}
\resizebox{\linewidth}{!}{
\begin{threeparttable}
\begin{tabular}{c|c|ccccc|ccccc|cc}
\toprule
\multirow{2}{*}{Method} & \multirow{2}{*}{Setting} & \multicolumn{5}{c|}{PETA} & \multicolumn{7}{c}{RAPv2} \\ \cline{3-14} 
	&	& mA & Accu & Prec & Recall & F1 & mA & Accu & Prec & Recall & F1 & Params(M) & MACs(G) \\ \hline \hline
\multirow{2}{*}{MsVAA\cite{sarafianos2018deep}\footnotemark[1]} & existing & 84.35 & 78.69 & 87.27 & 85.51 & 86.09 & 77.87 & 67.19 & 79.03 & 79.79 & 79.04 & \multirow{2}{*}{141.27} & \multirow{2}{*}{6.28}\\ \cline{2-2}
 & zero-shot & 71.03 & \bf{59.38} & 74.75 & 70.10 & 72.37 & 71.32 & 63.59 & 77.22 & 76.62 & 76.44\\\cline{1-2} \cline{13-14}
\multirow{2}{*}{VAC\cite{guo2019visual}\footnotemark[1]} & existing & 83.63 & 78.94 & 87.63 & 85.45 & 86.23 & 76.74 & 67.52 & 80.42 & 78.78 & 79.24 & \multirow{2}{*}{23.61} & \multirow{2}{*}{14.335} \\\cline{2-2}
 & zero-shot & 71.05 & 58.90 & 74.98 & 70.48 & 72.13 & 70.20 & \bf{65.45} & 79.87 & 76.65 & 77.07\\\cline{1-2} \cline{13-14}
 \multirow{2}{*}{ALM\cite{tang2019Improving}\footnotemark[1]} & existing & 84.24 & 77.84 & 85.79 & 85.60 & 85.41 & 78.21 & 66.98 & 78.25 & 80.43 & 78.93 & \multirow{2}{*}{30.86} & \multirow{2}{*}{4.32} \\\cline{2-2}
 & zero-shot & 70.67 & 58.56 & 72.97 & \bf{71.31} & 71.65 & 71.97 & 64.52 & 77.28 & \bf{77.74} & 77.06 \\\cline{1-2} \cline{13-14}
\multirow{2}{*}{Baseline} & existing & 85.11 & 79.14 & 86.99 & 86.33 & 86.39 & 77.34 & 66.12 & 81.99 & 75.62 & 78.21 & \multirow{2}{*}{23.61} & \multirow{2}{*}{4.05} \\\cline{2-2} 
 & zero-shot & 71.84 & 58.77 & \bf{77.06} & 68.24 & 71.72 & 70.83 & 63.63 & \bf{82.28} & 72.22 & 76.34\\ \bottomrule
\end{tabular}
\begin{tablenotes}
      \item \textbf{${}^\star$ Results are reimplemented with the same setting of our baseline for a fair comparison.}
\end{tablenotes}
\end{threeparttable}}
\vspace{-1em}
\label{tab:proposed_dataset_perf}
\end{table}

\begin{table}[h]
\centering
\caption{Performance comparison with state-of-the-art methods on the PETA, RAPv1, PA100k datasets. Five metrics, mA, Accuracy, Precision, Recall, F1 are evaluated.}
\resizebox{\linewidth}{!}{
\begin{tabular}{c|c|ccccc|ccccc|ccccc}
\toprule
\multirow{2}{*}{Method} & \multirow{2}{*}{Backbone} & \multicolumn{5}{c|}{PETA} & \multicolumn{5}{c|}{PA100k} &\multicolumn{5}{c}{RAPv1} \\ \cline{3-17} 
	&	& mA & Accu & Prec & Recall & F1 & mA & Accu & Prec & Recall & F1 & mA & Accu & Prec & Recall & F1 \\ \hline \hline
DeepMAR \cite{li2015deepmar} (ACPR15) & CaffeNet & 82.89 & 75.07 & 83.68 & 83.14 & 83.41 & 72.70 & 70.39 & 82.24 & 80.42 & 81.32 & 73.79 & 62.02 & 74.92 & 76.21 & 75.56 \\
HPNet\cite{liu2017hydraplus} (ICCV17) & InceptionNet & 81.77 & 76.13 & 84.92 & 83.24 & 84.07 & 74.21 & 72.19 & 82.97 & 82.09 & 82.53 & 76.12 & 65.39 & 77.33 & 78.79 & 78.05 \\
JRL \cite{wang2017attribute} (ICCV17) & AlexNet & 82.13 & -- & 82.55 & 82.12 & 82.02 &--&--&--&--&--& 74.74 & -- & 75.08 & 74.96 & 74.62 \\
LGNet \cite{liu2018localization} (BMVC18) & Inception-V2 &--&--&--&--&--& 76.96 & 75.55 & 86.99 & 83.17 & 85.04 & 78.68 & 68.00 & 80.36 & 79.82 & 80.09 \\
PGDM \cite{li2018pose} (ICME18) & CaffeNet & 82.97 & 78.08 & 86.86 & 84.68 & 85.76 & 74.95 & 73.08 & 84.36 & 82.24 & 83.29 & 74.31 & 64.57 & 78.86 & 75.90 & 77.35\\ \hline
MsVAA\cite{sarafianos2018deep}(ECCV18) & ResNet101 & 84.59 & 78.56 & 86.79 & 86.12 & 86.46 &--&--&--&--&--&--&--&--&--&--\\
VAC \cite{guo2019visual} (CVPR19) & ResNet50 &--&--&--&--&--& 79.16 & 79.44 & 88.97 & 86.26 & 87.59 &--&--&--&--&-- \\
ALM\cite{tang2019Improving} (ICCV19) & BN-Inception & 86.30 & 79.52 & 85.65 & 88.09 & 86.85 & 80.68 & 77.08 & 84.21 & 88.84 & 86.46 & 81.87 & 68.17 & 74.71 & 86.48 & 80.16  \\ 
\hline
FocalLoss & ResNet50 & 83.00 & 76.18 & 84.85 & 84.44 & 84.31 & 78.49 &  77.87  & 86.96 & 85.00 & 85.58 & 77.32 & 65.91 & 80.74 & 75.13 & 76.47 \\
Baseline & ResNet50 & 85.11 & 79.14 & 86.99 & 86.33 & 86.39 & 79.38, & 78.56 & 89.41 & 84.78 & 86.55 & 78.48 & 67.17 & 82.84 & 76.25 & 78.94 \\\bottomrule
\end{tabular}}
\vspace{-1.5em}
\label{tab:peta_rap_perf}
\end{table}

\noindent {\bf Results on existing datasets} Experiments are also conducted on existing PETA, RAPv1, PA100k datasets to make a comparison with recent methods and results are reported in Table~\ref{tab:peta_rap_perf}. We compare with state-of-the-art methods, including MsVAA\cite{sarafianos2018deep}, VAC \cite{guo2019visual} and ALM\cite{tang2019Improving}. According to experiments, we have following observations: 
1) Considering the overlapped identities in train and test set on existing PETA and RAPv1 datasets, performance in PA100k are more convinced.
3) Our proposed method with backbone ResNet50 achieves a better performance with only 16.71\% parameters and 64.49\% computation than  MsVAA method which adopts ResNet101 as backbone.
4) Compared to VAC \cite{guo2019visual} using extra training augmentation and two-branch network, our baseline method achieves a comparable performance with only 28.25\% computation in PA100k. 
 Replacing linear classifier with cosine-distance based classifier, our method obtains 86.02\%, 80.71\, 80.07\% mA in PETA, RAPv1, PA100K, which outperforms baseline by 0.91\%, 2.23\%, 0.69\% respectively. Consistent improvements achieved by proposed method compared to baseline demonstrate our strategy effectiveness, \textit{i.e.} normalizing the classifier weight of attributes to make it independent on positive samples of attributes.
 
% \footnotetext[1]{Results are achieved by the ensemble model.}
% \footnotemark[1]
%----------------------------------------------------------------------

\vspace{-1em}
\begin{table}[h]
\vspace{-1em}
\centering
\caption{Performance comparison with reimplement methods and their baseline on PETA, RAPv1, PA100k. Five metrics, mA, Accuracy, Precision, Recall, F1 are evaluated. Parameters(Params) and multiply-accumulate operations(MACs) of various methods are also reported.}
\resizebox{\linewidth}{!}{
\begin{threeparttable}
\begin{tabular}{c|c|ccccc|ccccc|ccccc|cc}
\toprule
\multirow{2}{*}{Method} & \multirow{2}{*}{Backbone} & \multicolumn{5}{c|}{PETA} & \multicolumn{5}{c|}{PA100k} &\multicolumn{5}{c}{RAPv1} \\ \cline{3-19} 
	&	& mA & Accu & Prec & Recall & F1 & mA & Accu & Prec & Recall & F1 & mA & Accu & Prec & Recall & F1 & Params(M) & MACs(G)\\ \hline
Baseline(MsVAA\cite{sarafianos2018deep}) & ResNet101 & 82.67 & 76.63 & 85.13 & 84.46 & 84.79 &--&--&--&--&--&--&--&--&--&--&--&--\\
Baseline(VAC\cite{guo2019visual}) & ResNet50 &--&--&--&--&--& 78.12 & 75.23 & 88.47 & 83.41 & 85.86 &--&--&--&--&--&--&-- \\
Baseline(ALM\cite{tang2019Improving}) & BN-Inception & 82.66 & 77.73 & 86.68 & 84.20 & 85.57 & 77.47 & 75.05 & 86.61 & \bf{85.34} & 85.97 & 75.76 & 65.57 & 78.92 & \bf{77.49} & 78.20 &--&-- \\ 
Baseline(ours) & ResNet50 & \bf{85.11} & \bf{79.14} & \bf{86.99} & \bf{86.33} & \bf{86.09} & \bf{79.38}, & \bf{78.56} & \bf{89.41} & 84.78 & \bf{86.25} & \bf{78.48} & \bf{67.17} & \bf{82.84} & 76.25 & \bf{78.94} &--&--\\
\hline
MsVAA\cite{sarafianos2018deep}(ECCV18)$^\star$ & ResNet50 & 84.35 & 78.69 & 87.27 & 85.51 & 86.09 & \bf{80.10} & 76.98 & 86.26 & 85.62 & 85.50 & 79.75 & 65.74 & 77.69 & 78.99 & 77.93 & 141.27 & 4.93\\
VAC \cite{guo2019visual} (CVPR19)$^\star$ & ResNet50 & 83.63 & 78.94 & \bf{87.63} & 85.45 & 86.23 & 79.04 & 78.25 & 88.01 & \bf{86.07} & \bf{86.83} & 78.47 & \bf{68.55} & 81.05 & \bf{79.79} & \bf{80.02} & 23.61 & 14.335\\ 
ALM\cite{tang2019Improving} (ICCV19)$^\star$ & ResNet50 & 84.24 & 77.84 & 85.79 & 85.60 & 85.41 & 77.47 & 75.05 & 86.61 & 85.34 & 85.97 & 75.76 & 65.57 & 78.92 & 77.49 & 78.20 & 30.86 & 4.32\\\bottomrule
\end{tabular}
\begin{tablenotes}
      \item \textbf{${}^\star$ Results are reimplemented with the same setting of our baseline for a fair comparison..}
\end{tablenotes}
\end{threeparttable}
}
\vspace{-4em}
\label{tab:baseline}
\end{table}

\subsection{Ablation Study of Baseline}
To make a fair comparison with previous SOTA methods, we reimplement MsVAA, VAC, ALM methods and report their performance on PETA, PA100k, RAPv1 as well as their corresponding baseline performance, as shown in Table.~\ref{tab:baseline}. It is worth noting that our baseline achieve a much better performance than baseline of previous methods, even if they adopt a powerful backbone ResNet101 \cite{he2016deep}. And compared to previous SOTA methods reimplemented with same backbone, our baseline achieve a comparable even better performance. We argue that the effectiveness of a method can not be fully verified when compared with a worse baseline. 

\begin{figure}
\vspace{-1em}
\centering
\includegraphics[width=0.8\linewidth]{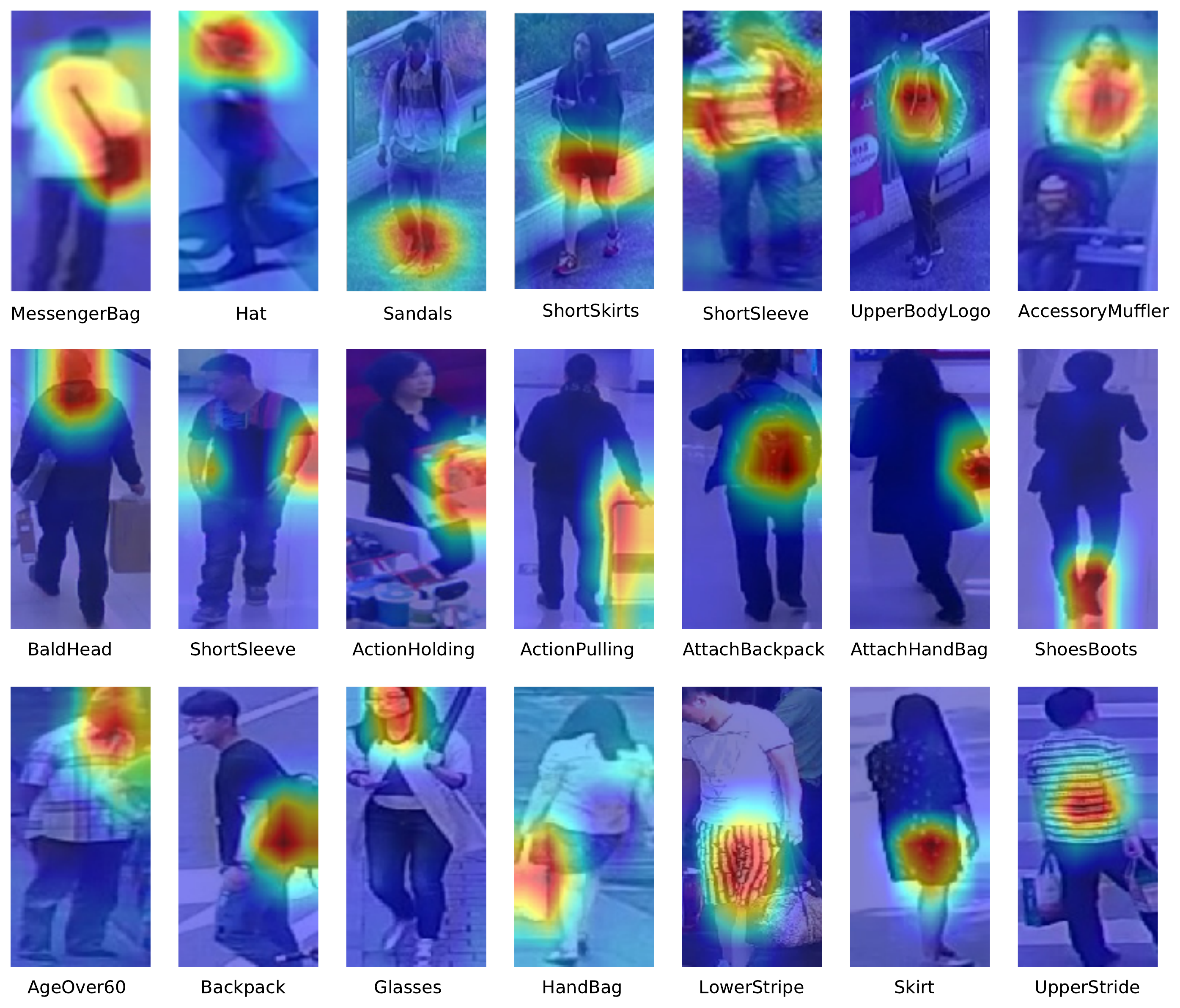}
\caption{Specific-attribute attention area of our baseline method on PETA, RAP, PA100k datasets from top to down.}
\label{fig:attention}
\vspace{-1em}
\end{figure}

The reason why our baseline can achieve a comparable even better performance than previous methods is that a strong baseline itself can implicitly learn the location of attribute-specific area. We utilize GradCAM \cite{selvaraju2017grad} to locate discriminative visual cues of our baseline model. As show in Fig.~\ref{fig:attention}, even if without explicitly modeling the localization of attribute-specific area, our baseline can localize attribute-specific area to learn discriminative representation. The important thing is not to locate the area of specific attribute, but to distinguish the fine-grained attributes in the same are, such as  distinguish sandals from sneakers. Compared to original performance of SOTA methods, there is little difference in performance of reimplemented methods except for ALM. The reason is that attention area of ALM is hard bounding-box, which is coarse-grained and introduces environmental noise.

\section{Conclusion} \label{conclusion}
In this paper, we propose two realistic datasets PETA\textsubscript{$zs$} and RAPv2\textsubscript{$zs$} to solve the unreasonable and impractical setting of existing datasets, which misleads model evaluation. Meanwhile, we find SOTA methods can not make any performance improvement on our strong baseline.

\clearpage
\bibliographystyle{splncs04}
\bibliography{jian.bib}
\end{document}